\documentclass[runningheads]{llncs}

\usepackage{eccv}

\usepackage{eccvabbrv}

\usepackage[T1]{fontenc}

\DeclareFontFamily{T1}{lmr2}{}
\DeclareFontShape{T1}{lmr2}{m}{n}{<->ec-lmr12}{}
\DeclareFontShape{T1}{lmr2}{m}{sl}{<->ec-lmro12}{}
\DeclareFontShape{T1}{lmr2}{m}{it}{<->ec-lmri12}{}
\DeclareFontShape{T1}{lmr2}{m}{sc}{<->ec-lmcsc10}{}
\DeclareFontShape{T1}{lmr2}{bx}{n}{<->ec-lmbx12}{}
\DeclareFontShape{T1}{lmr2}{bx}{it}{<->ec-lmbxi10}{}
\DeclareFontShape{T1}{lmr2}{bx}{sl}{<->ec-lmbxo10}{}
\DeclareFontShape{T1}{lmr2}{bx}{sc}{<->ssub*cmr/bx/sc}{}

\usepackage{microtype}

\usepackage{algorithm}
\usepackage{algpseudocode}
\usepackage{amsfonts}
\usepackage{amsmath}
\usepackage{booktabs}
\usepackage{enumitem}
\usepackage{graphicx}
\usepackage{hhline}
\usepackage{mathtools}
\usepackage{multirow}
\usepackage{nicefrac}
\usepackage{pdfpages}
\usepackage{pifont}
\usepackage{soul}
\usepackage{tabularx}
\usepackage{url}
\usepackage{wrapfig}
\usepackage{xcolor}

\usepackage[accsupp]{axessibility}  % Improves PDF readability for those with disabilities.

\usepackage{hyperref}

\usepackage{orcidlink}

\usepackage{amsmath,amsfonts,bm}

\def\eqref#1{equation~\ref{#1}}
\def\1{\bm{1}}

\DeclareMathAlphabet{\mathsfit}{\encodingdefault}{\sfdefault}{m}{sl}
\SetMathAlphabet{\mathsfit}{bold}{\encodingdefault}{\sfdefault}{bx}{n}

\def\gN{{\mathcal{N}}}

\newcommand{\E}{\mathbb{E}}

\newcommand{\methodname}{\mbox{\textsc{{{Videoshop}}}}}
\newcommand{\methodnameshort}{\mbox{\textsc{{{Videoshop}}}}}

\newsavebox\CBox
\def\textBF#1{\sbox\CBox{#1}\resizebox{\wd\CBox}{\ht\CBox}{\textbf{#1}}}

\newcommand{\cskip}{c_{\text{skip}}}
\newcommand{\cin}{c_{\text{in}}}
\newcommand{\cout}{c_{\text{out}}}
\newcommand{\cnoise}{c_{\text{noise}}}

\newcommand*\circled[1]{\tikz[baseline=(char.base)]{
    \node[shape=circle,draw,inner sep=0.5pt] (char) {\small #1};}}

\newcommand*\samethanks[1][\value{footnote}]{\footnotemark[#1]}

\begin{document}

% \title{Video Point \& Edit: Precise Video Editing with Noise-Extrapolated Diffusion Inversion}
\title{\parbox{0.05\textwidth}{\includegraphics[width=\linewidth]{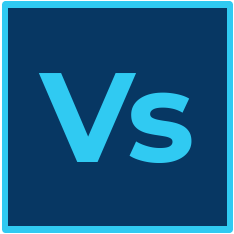}} \textsc{\methodname}: Localized Semantic Video Editing with Noise-Extrapolated Diffusion Inversion}

\titlerunning{\textsc{\methodname}}

\author{Xiang Fan\inst{1} \and
Anand Bhattad\thanks{equal advising}\inst{2} \and
Ranjay Krishna\samethanks\inst{1}}

\authorrunning{X.~Fan et al.}

\institute{University of Washington \and
Toyota Technological Institute at Chicago
% \email{lncs@springer.com}
\\
\url{https://videoshop-editing.github.io/} 
% \and
}

\maketitle

\begin{figure}
  \vspace{-2em}
  \centering
  \includegraphics[width=\textwidth]{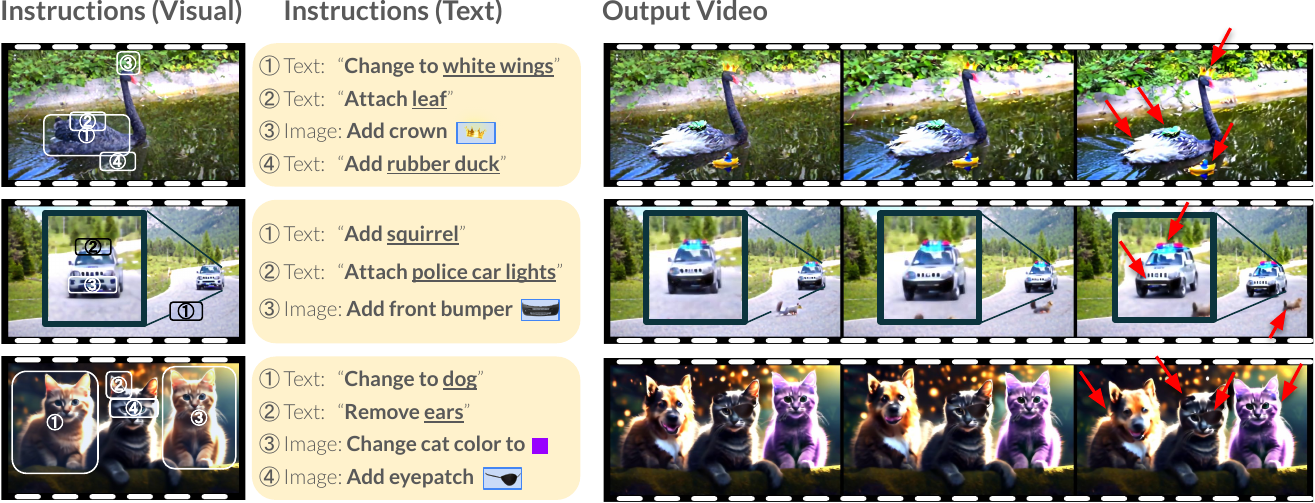}
  \vspace{-12pt}
  \caption{\methodname\ is a training-free method for precise video editing. 
  Given an original video and user edits to the first frame, \methodname\ automatically propagates the changes to all the frames of the video while maintaining semantic, geometric, and temporal consistency.
  To edit the first frame, users can leverage image editing tools, including text-based inpainting and professional editing software like Photoshop. As such, \methodname\ supports video edits congruent with possible image edits that Photoshop enables: users can add new objects, remove objects or their parts, modify attributes, etc.}
  \label{fig:teaser}
\end{figure}

\vspace{-3.2em}
\begin{abstract}
  We introduce \methodname, a training-free video editing algorithm for localized semantic edits.
\methodname\ allows users to use any editing software, including Photoshop and generative inpainting, to modify the first frame; it automatically propagates those changes, with semantic, spatial, and temporally consistent motion, to the remaining frames.
Unlike existing methods that enable edits only through imprecise textual instructions, \methodname\ allows users to add or remove objects, semantically change objects, insert stock photos into videos, etc.\ with fine-grained control over locations and appearance.
We achieve this through image-based video editing by inverting latents with noise extrapolation, from which we generate videos conditioned on the edited image.
\methodname\ produces higher quality edits against 6 baselines on 2 editing benchmarks using 10 evaluation metrics.


\end{abstract}

\vspace{-2.2em}
\section{Introduction}
\vspace{-.6em}

Traditional video editing requires sophisticated direct manipulation and is a manually exhaustive process~\cite{mackay1994video,santosa2013direct}. Software tools like Adobe Premiere and Apple Final Cut have a steep learning curve and are still limited in their ease of use~\cite{AdobePremierePro,AppleFinalCutPro}. While they support operations like stitching together video clips, they provide little to no support for \textit{propagating edit changes} from one frame to another. Consider wanting to change the wings of a swan to have white features, add a crown on top of the swan's head, or even add a rubber duck wading along with the swan (\cref{fig:teaser}). Such edits are only possible with painstaking per-frame manual curation.

Current video models fall short in facilitating precise, localized semantic editing. Open-source video diffusion models~\cite{blattmann2023stable} have spurred new algorithmic developments~\cite{kara2023rave,ceylan2023pix2video,qi2023fatezero,yatim2023spacetime}, yet they do not support the finesse required for localized edits. Models typically require extensive fine-tuning for individual videos or rely on coarse textual instructions that lack specificity~\cite{zuo2023cutandpaste,shin2023editavideo,geyer2023tokenflow}. Moreover, they struggle to introduce new objects with independent motions~\cite{yatim2023spacetime,kahatapitiya2024objectcentric}.

Image editing has improved control over still frames~\cite{zhang2023adding}, but ensuring temporal consistency across video frames remains unresolved. Frame-by-frame adjustments do not guarantee the continuity essential for coherent video sequences, revealing a gap between static image manipulation and dynamic video editing.

To address this limitation, we introduce \methodname, a training-free video editing algorithm that enables users to make localized semantic edits. \methodname\ allows users to make any modification they want to the first frame of the video; it propagates those changes, with temporally consistent motion, to all the remaining frames. 
The changes to the first frame can be made using any image editing tool~\cite{michel2024object, bhattad2022cut, goel2023pair, chen2023anydoor, zhang2023magicbrush, yenphraphai2024image, hertz2022prompt, bhattad2024stylitgan}. Users can use image editing models like ControlNet~\cite{zhang2023adding} or Text Inversion~\cite{gal2022image}; they can even load the frame onto Adobe Photoshop to edit pixels manually, add clip art, use content-aware fill, or any of Photoshop's features.
\methodname\ does not require finetuning and leverages Stable Video Diffusion\cite{blattmann2023stable} to make the edits.
In other words, \methodname\ can edit 14-frame videos within an average of 2 minutes, and supports edits within the large domain of videos.
As video diffusion models themselves improve to support longer videos, \methodname\ will be able to edit even longer videos.

Two technical insights enable \methodname: (1) noticing that video latents follow a near-linear trajectory during the denoising process, and that (2) the VAE encoder is unnormalized, resulting in a high variance in the magnitude of the latents.
To contextualize these insights, let's review how traditional image inversion enables image editing.
Input images are inverted using a denoising diffusion implicit model (DDIM) to generate intermediate latents across time steps deterministically.
This process approximates the noise at time step $t+1$ using the latent from time step $t$. We find that this approximation reconstructs only the first frame accurately in image-to-video diffusion, resulting in unstable latents due to cumulative approximation errors. (1) Our investigations reveal that the latents are near-linear during the denoising process. We capitalize on this observation and introduce \textbf{inversion with noise extrapolation}, a mechanism for achieving faithful reconstruction for any video.
(2) Our investigations also reveal that the video latents are unnormalized, leading to further instability. We introduce a \textbf{latent normalization} technique to ensure consistency and quality.

Our extensive experiments show that \methodname\ produces higher quality edits against $6$ baselines on $2$ editing benchmarks using $10$ evaluation metrics. This method empowers users to make direct pixel modifications, enabling a spectrum of semantic edits (refer to \cref{fig:teaser}). Examples include ``transforming a specific cat into a dog'', ``removing the ears of a cat'', ``changing a cat's fur color to purple'', ``adding dynamic objects such as a squirrel on the road'' or ``police car lights'' that respond naturally to the car's movements. {\methodname} not only ensures geometric fidelity where the car lights appropriately turn with the vehicle but also captures realistic motion, as exemplified by a rubber duck floating in unison with the swan and a squirrel that scurries away as a police car draws near.

On a broader scale, {\methodname} equips users with video manipulation capabilities akin to those provided by image editing software like Photoshop, enabling new potential applications that previously would have been prohibitively challenging.

\vspace{-.5em}
\section{Related Work}
\vspace{-.5em}

\noindent\textbf{Image and video generation.} 
Generative Adversarial Networks (GANs)~\cite{goodfellow2014generative} initially advanced the quality of image generation, with notable architectures such as StyleGAN~\cite{karras2019style}. Despite the community's efforts, GANs remain difficult to train, limiting their ability to generate videos ~\cite{kang2023scaling,sauer2023stylegan}. 
Diffusion models, including Denoising Score Matching~\cite{vincent2011connection} and Noise-Contrastive Estimation~\cite{gutmann2010noise}, are much easier to scale~\cite{karras2022elucidating,ho2020denoising,rombach2022highresolution}. 
Today's video generation models are generally diffusion models~\cite{singer2022make, blattmann2023align, blattmann2023stable, kondratyuk2023videopoet, menapace2024snap,Mullan_Hotshot-XL_2023,wang2023modelscope,chen2024videocrafter2,wang2023videocomposer,zhang2023show1}.
We develop our technique using Stable Video Diffusion (SVD)~\cite{blattmann2023stable}, which is based on the EDM~\cite{karras2022elucidating} framework and generates high-quality videos conditioned on a first frame.

\vspace{.8em}
\noindent\textbf{Text-based video editing.} Current video editing methods are commonly text-based, which modify video based on textual instructions. Such tasks include motion transfer~\cite{yatim2023spacetime,chan2019everybody,yin2023dance}, object editing~\cite{qi2023fatezero,kahatapitiya2024objectcentric,kara2023rave,shin2023editavideo,ren2024customizeavideo,zuo2023cutandpaste}, attribute editing~\cite{kara2023rave,jeong2024groundavideo}, texture editing~\cite{couairon2023videdit,kasten2021layered}, and style editing~\cite{qi2023fatezero,kahatapitiya2024objectcentric,ceylan2023pix2video,geyer2023tokenflow,yang2023rerender}. 
A common approach is to use a text-to-video model~\cite{wang2023modelscope} and control the denoising process such that the generated result satisfies the editing conditions. However, textual instructions alone provide minimal specification, limiting most practical applications beyond modifying object classes, attributes, and textures.

\vspace{.8em}
\noindent\textbf{DDIM inversion.} Denoising Diffusion Implicit Models (DDIMs)~\cite{song2022denoising} are a class of diffusion models that deterministically generate samples from a random noise. 
Due to its determinism, DDIM models can be inverted to find the corresponding noise given a sample. In the context of image editing, several methods have been proposed to improve the quality the edited images, including null-text inversion~\cite{mokady2022nulltext} and mathematically-exact inversion methods~\cite{wallace2022edict,meiri2023fixedpoint,zhang2023exact}. 
EDICT~\cite{wallace2022edict} and BDIM~\cite{zhang2023exact} propose an alternative denoising process that tracks more than one latent variables to derive an exact solution to the inversion formula. Another proposes a fixed-point iteration method to solve the inversion formula~\cite{meiri2023fixedpoint}.
Unfortunately, applying diffusion inversion to text-to-image models can lead to a loss of temporal consistency in the resulting video~\cite{qi2023fatezero}.

\vspace{.8em}
\noindent\textbf{Layered video editing.} Another approach to video editing is through layered atlases, which decompose video frames into several layers (often foregrounds and backgrounds) and edit the layers corresponding to a target (\eg NLA~\cite{kasten2021layered} and DiffusionAtlas~\cite{chang2023diffusionatlas}).
However, atlases do not generate new motions or support edits beyond changes to object class, attribute, and texture~\cite{chang2023diffusionatlas}.

\vspace{-.6em}
\section{Method}
\vspace{-.6em}

Given an input video $\mathcal{V}$ and an edited first frame $\mathcal{I}$, our goal
is to generate a new video $\mathcal{J}$ that preserves the overall motion and semantics of the original video $\mathcal{V}$, while propagating the changes made to the first frame $\mathcal{I}$. Before we explain our method, we revisit diffusion models, DDIM inversion, and the EDM framework.

\vspace{-.5em}
\subsection{Background on latent diffusion models}

\textbf{Denoising Diffusion Probabilistic Models} (DDPM)~\cite{ho2020denoising} are a class of models that generate an image or video from a random Gaussian noise through a sequence of $T$ denoising steps. The denoising process is defined as a sequence of timesteps $T, T - 1, ..., 0$. At timestep $T$, a $F\times H\times W\times 3$-dimensional random noise is sampled from a multivariate normal distribution, denoted as the initial noise $x_T$. DDPM applies a denoising neural network to iteratively de-noise latents $x_{t+1} \rightarrow x_t$ until $x_0$. The final $x_0$ is a generated $F\times H\times W\times 3$-dimensional video with $F$ frames, and spatial size of $H\times W$.

Denoising in pixel space is computationally expensive as each denoising step needs to produce a $F\times H\times W\times 3$-dimensional noise estimation. \textbf{Latent Diffusion Models}~\cite{rombach2022highresolution} instead encode the sample image into a much smaller latent space using a pretrained variational autoencoder (VAE) with a $F\times h\times w\times c$-dimensional hidden representation, such that $F\times h\times w\times c \ll F\times H\times W\times 3$.
Latent models sample $x_T$ in the latent space and iteratively denoise until $x_0$.
Finally, $x_0$ is decoded using the pretrained VAE decoder into the pixel space.

Latent models are trained by first sampling a video encoded as $x_0$, a timestep $t$, and a random noise $\epsilon$. From these, a noise $\epsilon_t$ corresponding to the timestep $t$ is calculated and added to $x_0$ to produce $x_t$.
A denoising, timestep-conditioned U-Net $\epsilon_\theta$, is trained to estimate $\epsilon_t$ given $x_t$:
\begin{align}
  \fontfamily{lmr2}\selectfont
  \E_{x,t,\epsilon \sim \gN(0, 1)}\lVert \epsilon_t - \epsilon_\theta(x_t; t) \rVert^2
\end{align}

\vspace{-1.5em}
\subsection{Background on diffusion inversion}

\textbf{Denoising Diffusion Implicit Models} (DDIMs)~\cite{song2022denoising} are a class of latent diffusion models with a deterministic denoising process. DDIMs deterministically generate samples from a random noise when its denoising step noise parameter is set to zero. This determinism allows for the diffusion inversion process. 

The \textbf{inversion process} is a technique used to generate the deterministic latents $x_T$ given a video $\mathcal{V}$. From $\mathcal{V}$, we first extract $x_0$ from the pretrained variational autoencoder's encoder.
This process calculates the latents $\hat{x}_t$ for $t=1\ldots T$, given $x_0$.
If the inversion process is accurate, denoising $\epsilon_\theta(x_t; t)$ iteratively starting from $\hat{x}_T$ should yield $\hat{x}_0$, such that $\hat{x}_0 \approx x_0$.

In \textbf{diffusion-based image editing}, it is common to first invert the image latents $x_0$ into its corresponding noise $\hat{x}_T$, and apply the denoising process with a modified conditioning text to obtain the edited image~\cite{mokady2022nulltext}. 
Common image diffusion models are conditioned on a text prompt~\cite{rombach2022highresolution}.

With deterministic sampling, the denoising step from $x_{t+1} \rightarrow x_t$ is defined as:
\vspace{-.8em}
\begin{align}
  x_{t} &= \sqrt{\alpha_{t}} \underbrace{\left( \frac{x_{t+1} - \sqrt{1 - \alpha_{t+1}} \epsilon_\theta(x_{t+1}; t+1)}{\sqrt{\alpha_{t+1}}} \right)}_{\text{\fontfamily{lmr2}\selectfont predicted }x_0} + \underbrace{\sqrt{1 - \alpha_{t}} \cdot \epsilon_\theta(x_{t+1}; t+1)}_{\text{\fontfamily{lmr2}\selectfont direction pointing to }x_{t+1}}
  \text{\vspace{-4em}}
\end{align}
where $\epsilon_\theta$ is the noise-predicting U-Net and $\alpha_t$ is the noise scheduling parameter.

\textbf{Stable Video Diffusion}~\cite{blattmann2023stable} is a video diffusion model that conditions on the first-frame image instead of text. 
Ideally, one would invert the video latents similar to the inversion process mentioned above and apply edits by conditioning on an edited first frame.
However, we observe that naively inverting the video latents often results in an incoherent video (\cref{fig:ablation_examples}), necessitating a new method.

\subsection{Background on the EDM framework}

Stable Video Diffusion~\cite{blattmann2023stable} employs the \textbf{EDM framework}~\cite{karras2022elucidating}, which improves upon DDIM with a reparameterization to the denoising process.
A denoising step from $x_{t+1} \rightarrow x_t$ of deterministic sampling in the EDM framework is:
\begin{align}
  x_t &= x_{t+1} + \underbrace{\frac{\sigma_t - \sigma_{t+1}}{\sigma_{t+1}} \left( x_{t+1} - \overbrace{\left( \cskip^{t+1} x_{t+1} + \cout^{t+1} F_\theta \left( \cin^{t+1} x_{t+1}; \cnoise^{t+1} \right) \right)}^{\text{\fontfamily{lmr2}\selectfont predicted }x_0} \right)}_{\text{\fontfamily{lmr2}\selectfont noise removed at step }t} \label{eq:edm_forward}
\end{align}
where $\sigma_t$ is the scheduled noise level at step $t$. $\cskip^{t}$, $\cin^{t}$, $\cout^{t}$, and $\cnoise^{t}$ are coefficients dependent on the noise schedule and the current step $t$. $F_\theta$ is a neural network parametrized by $\theta$.

To perform \textbf{inversion in the EDM Framework,} 
we can rewrite \cref{eq:edm_forward} as an inversion step $\hat{x}_{t} \rightarrow \hat{x}_{t+1}$:
\begin{align}
  \hat{x}_{t+1} &= \frac{\sigma_{t+1}\hat{x}_t + \left( \sigma_t - \sigma_{t + 1} \right) \cout^{t+1} F_\theta \left( \cin^{t+1} \hat{x}_{t+1}; \cnoise^{t+1} \right)}{\left( \sigma_t - \sigma_{t+1} \right) \left( 1 - \cskip^{t+1} \right) + \sigma_{t+1}} \label{eq:edm_backward}
\end{align}
However, because the input to $F_\theta$ is dependent on the next inverted latent $\hat{x}_{t+1}$, \cref{eq:edm_backward} is not directly solvable. Naive inversion methods approximate $\hat{x}_{t+1}$ with $\hat{x}_t$ such that:
\begin{align}
  F_\theta \left( \cin^{t+1} \hat{x}_{t+1}; \cnoise^{t+1} \right) \approx F_\theta \left( \cin^{t} \hat{x}_{t}; \cnoise^{t+1} \right) \label{eq:naive_inversion}
\end{align}

\subsection{Limitations with naive video inversion to EDM}

While common in inverting image diffusion models~\cite{mokady2022nulltext,ceylan2023pix2video}, naive inversion (\cref{eq:naive_inversion}) leads to latents that only correctly reconstruct the first video frame, when directly applied to Stable Video Diffusion. We find that naive inversion (\cref{eq:naive_inversion}) introduces a compounding approximation error that we show quantitatively and qualitatively in \cref{sec:ablation}.

Several methods have been proposed to address this in image diffusion models~\cite{wallace2022edict,meiri2023fixedpoint,zhang2023exact}. Amongst them, EDICT~\cite{wallace2022edict} and FPI \cite{meiri2023fixedpoint} require additional passes through the model at each step, increasing computation cost. 
While BDIA~\cite{zhang2023exact} improves upon EDICT~\cite{wallace2022edict} by eliminating additional passes, both of them modify the denoising process. We find that BDIA destabilizes the latents in the inversion process and results in undesirable artifacts in the resulting video (\cref{fig:vs_baselines}).

\begin{figure}[t]
  \centering
  \includegraphics[width=\textwidth]{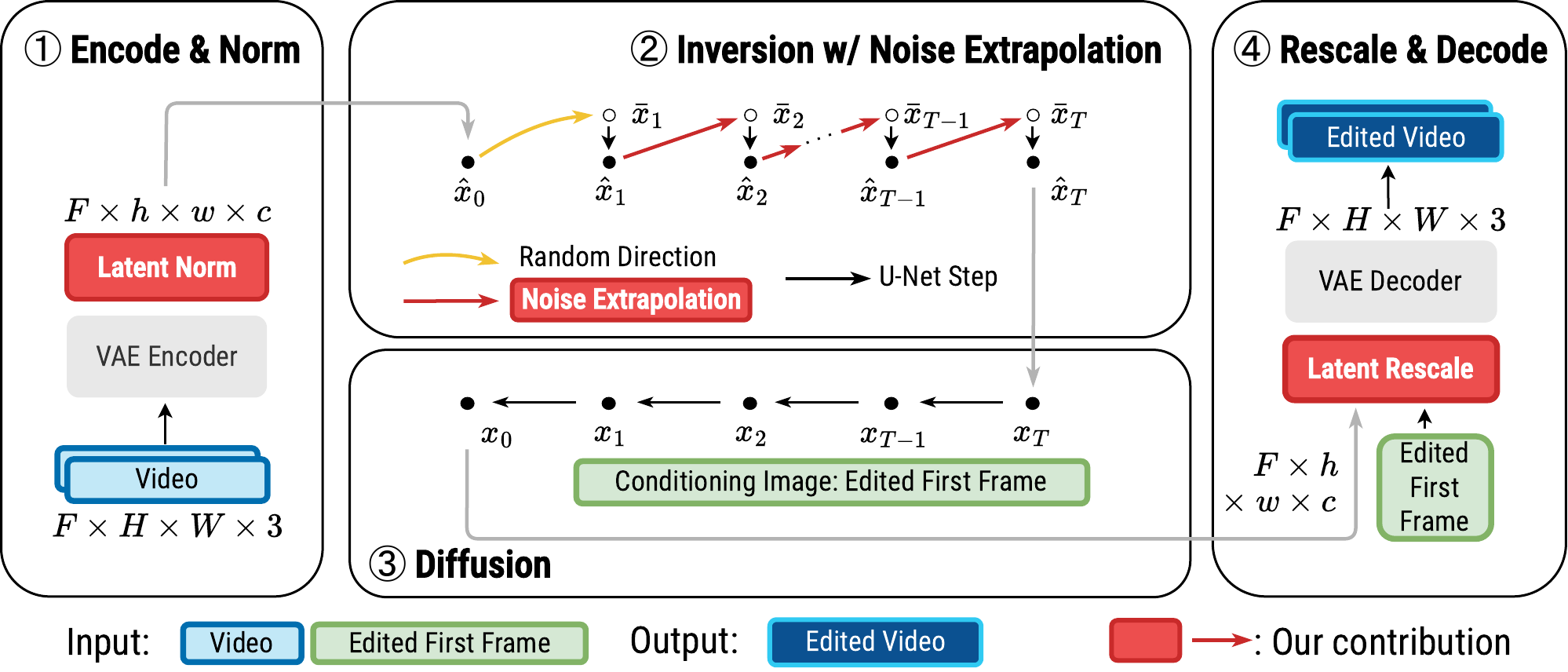}
    \vspace{-10pt}
  \caption{Overview of \methodname\ for localized semantic video editing. Our contributions are highlighted with red boxes and arrows. Our method includes four primary stages: (1) \textbf{Encode \& Norm}, where the input video is encoded into a latent space using a VAE encoder, followed by normalization to ensure stability throughout inversion. (2) In the \textbf{Inversion w/ Noise Extrapolation} phase, noise extrapolation is systematically applied at each step to provide a corrective term that guides the inversion trajectory, ensuring the video is mapped to correct latent noise. This step is key for aligning the latent space trajectory at every timestep. (3) \textbf{Diffusion} then ensures user edits are seamlessly integrated across the video sequence, enforcing consistency while diffusing the initial modifications through time. (4) The last step is \textbf{Rescale \& Decode}, where the now-edited latent sequence is rescaled to align with the original data's statistical distribution and decoded back into the video, resulting in an output video that reflects the desired semantic edits while maintaining the natural flow of the original sequence.}
  \label{fig:method}
  \vspace{-5pt}
\end{figure}

\subsection{Our contribution: Inverting with noise extrapolation}
\label{sec:noise_extrapolation}

Our goal is to find a better approximation for $F_\theta \left( \cin^{t+1} \hat{x}_{t+1}; \cnoise^{t+1} \right)$ in \cref{eq:edm_backward}. First, we observe that the latents in the denoising process maintain a near-linear trajectory. We then propose noise extrapolation to exploit this observation.

\begin{wrapfigure}{r}{.44\textwidth}
  \vspace{-0em}
  \centering
  \includegraphics[width=\linewidth]{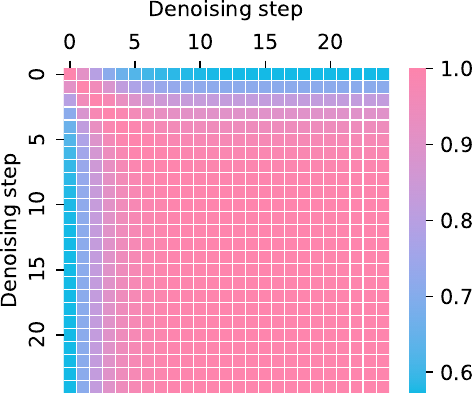}
  \vspace{-2em}
  \caption{Cosine similarity matrix for pairs of latent vectors throughout the denoising process. The latent vectors are approximately collinear, which supports our linear noise extrapolation.}
  \vspace{-5em}
  \label{fig:latent_analysis}
\end{wrapfigure}
\subsubsection{Near-linearity of $x_t$ trajectory.} It has been observed that the denoising trajectory of $x_t$ in image diffusion models is approximately linear~\cite{karras2022elucidating} at low and high noise levels. To investigate the latent trajectory of video diffusion models, we measure the average cosine similarity between vectors $x_t \rightarrow x_0$ and $x_{t'} \rightarrow x_0$ for pairs of $t, t'$ in the denoising process of $100$ random videos and show the results in \cref{fig:latent_analysis}. We observe high cosine similarities throughout the denoising process after the initial steps, suggesting that the trajectory of $x_t$ is approximately linear after the low noise levels, which we exploit in our noise extrapolation method \\ to invert $\hat{x}_t$.

\subsubsection{Quantifying the near-linear trajectory.} We measure the cosine similarities throughout the denoising process and report our results. The cosine similarity of $x_t - x_0$ and $x_{t'} - x_0$, averaged over all pairs of timesteps $t, t'$ and 100 video samples, is $\mathbf{0.9282}$. The cosine similarity of $x_t - x_0$ and $x_{t+1} - x_0$, averaged over timesteps $t$ and 100 video samples, is $\mathbf{0.9919}$. The minimum cosine similarity of $x_t - x_0$ and $x_{t+1} - x_0$, over all timesteps $t$ and 100 video samples, is $\mathbf{0.9107}$.

\vspace{-.7em}
\subsubsection{Noise extrapolation.} With this insight, we linearly extrapolate the noise at $\hat{x}_t$ to obtain an approximation of $\bar{x}_{t+1}$ to provide to $F_\theta$, as follows:
\vspace{-.6em}
\begin{align}
  \bar{x}_{t+1} &\approx
  \begin{cases}
  \underbrace{\frac{\sigma_{t+1}}{\sigma_t} \overbrace{(\hat{x}_t - x_0)}^{\sim \gN(0, \sigma_t)}}_{\sim \gN(0, \sigma_{t+1})} + x_0 &\qquad (\sigma_t > \Sigma) \vspace{1em} \\
  \gN(0, \sigma_{t+1}) + x_0 &\qquad (\sigma_t \leq \Sigma)
  \end{cases}
  \label{eq:noise_extrapolation}
\end{align}
\vspace{-1em}

\noindent
where $\Sigma$ is a threshold noise level. The threshold $\Sigma$ is necessary because at low $\sigma_t$, dividing by a small number results in a large noise, which can destabilize the latents. We ablate noise threshold in \cref{sec:ablation}.

Putting it together, we modify \cref{eq:edm_backward} to estimate the inversion using $\bar{x}_{t+1}$:
\vspace{-1.5em}
\begin{align}
  \hat{x}_{t+1} &= \frac{\sigma_{t+1}\hat{x}_t + \left( \sigma_t - \sigma_{t + 1} \right) \cout^{t+1} F_\theta \left( \cin^{t+1} \bar{x}_{t+1}; \cnoise^{t+1} \right)}{\left( \sigma_t - \sigma_{t+1} \right) \left( 1 - \cskip^{t+1} \right) + \sigma_{t+1}} \label{eq:edm_backward_with_noise_extrapolation}
\end{align}
\vspace{-1.2em}

\noindent
After obtaining the final $\hat{x}_T$ from the inversion process, we apply the denoising process to $\hat{x}_T$ conditioned on the edited image $\mathcal{I}$ to obtain the edited latents $x_0$.

\vspace{-.8em}
\subsection{Our contribution: Latent normalization and rescaling}
\vspace{-.3em}
\label{sec:latent_normalization} 
Direct output from the VAE encoder is unnormalized, resulting in a large variance in the magnitude of the final latent from the inversion process. We observe that this leads to poor quality in generated videos. To address this, we propose to normalize the latents before the start of the inversion process to unit standard deviation to stabilize the latents. After denoising, we rescale the latents with the mean and standard deviation of the latents of the target image:
\vspace{-.6em}
\begin{align}
  \hat{x}_0 = \frac{x_{\text{in}}}{\sigma_{\text{in}}} \qquad \qquad x_{\text{out}} = \frac{\sigma_{\text{img}}}{\sigma_{0}} \left(x_0 - \mu_{0} + \mu_{\text{img}}\right)
\end{align}
\vspace{-1.2em}

\noindent
where $x_\text{in}$ is the VAE-encoded latent sample, $\hat{x}_0$ is the input to the inversion process, $x_0$ is the output of the denoising process, and $x_\text{out}$ is the final output to be decoded by the VAE into video $\mathcal{J}$. $\mu_\text{img}$ and $\sigma_\text{img}$ are the mean and standard deviation of the VAE-encoded latents of the target image. All normalization are done per-channel. $\sigma_\text{in}$ is calculated across all frames; $\mu_0$, $\sigma_0$, $\mu_\text{img}$, and $\sigma_\text{img}$ are calculated for the first frame (the only available frame in the target image).

\vspace{-.6em}
\section{Experiments and Results}
\label{sec:experiments}
\vspace{-.2em}

The key takeaways from our experiments are as follows:
(1) \methodname\ successfully performs localized semantic video editing among a diverse set of edit types.
(2) Compared to \methodname, existing methods demonstrate clear limitations in maintaining visual fidelity to the source video and target edit.
(3) \methodname\ achieves SOTA performance in localized editing, as evaluated by edit fidelity and source faithfulness, while maintaining high temporal consistency. 
(4) In our user study, \methodname\ consistently outperforms text-based video editing methods, while maintaining high video generation quality. 
(5) \methodname\ is efficient, with an average speedup of 2.23x compared to the baselines.

\vspace{-.6em}
\subsubsection{Datasets.} We utilize two datasets for our experiments: a large-scale generated video editing dataset from the MagicBrush~\cite{zhang2023magicbrush} image-editing dataset and an expert-curated video editing dataset with source videos from HD-VILA-100M~\cite{xue2022advancing}.

The MagicBrush dataset is a manually annotated image editing dataset that contains over 10,000 tuples of ``\texttt{(source image, instruction, edit mask, edited image)}''. These tuples cover a wide range of edit types, including object addition, replacement, removal, and changes in action, color, texture, and counting. To convert MagicBrush into a video dataset, we generate videos conditioned on the source images using a video generation model\cite{blattmann2023stable}. The first frame of each generated video is conditioned to match the corresponding source image. The resulting dataset consists of ``\texttt{(source \underline{video}, instruction, edit mask, edited image)}'' tuples.

\begin{wraptable}{r}{.3\textwidth}
  \vspace{-4.4em}
  \caption{Types of edit in the expert dataset.}
  \label{tab:expert_dataset_breakdown}
  \centering
  \scriptsize
  \fontfamily{lmr2}\selectfont
  \begin{tabularx}{.3\textwidth}{l | *{1}{>{\centering\arraybackslash}X}}
    \toprule
    \textBF{Type of Edit} & \textBF{Percentage \%} \\
    \midrule
    \textBF{Add object} & 36\% \\
    \textBF{Change appearance} & 20\% \\
    \textBF{Remove object} & 18\% \\
    \textBF{Replace object} & 16\% \\
    \textBF{Change action} & \hphantom{0}6\% \\
    \textBF{Change color} & \hphantom{0}4\% \\
    \bottomrule
  \end{tabularx}
  \vspace{-4em}
\end{wraptable}
The HD-VILA-100M dataset is a large-scale high-resolution video dataset collected from YouTube, encompassing diverse open domains. From this dataset, we sample 45 videos and ask editing experts to provide edits on the first frame of each video, along with the corresponding edit instructions. The resulting expert dataset consists of ``\texttt{(source video, instruction, edit mask, edited image)}'' tuples. A summary of the edit types from the expert dataset is shown in \cref{tab:expert_dataset_breakdown}. All videos are resized to 14 frames with an aspect ratio of 16:9.

\begin{figure}[p]
  \centering
  \includegraphics[height=.9\textheight]{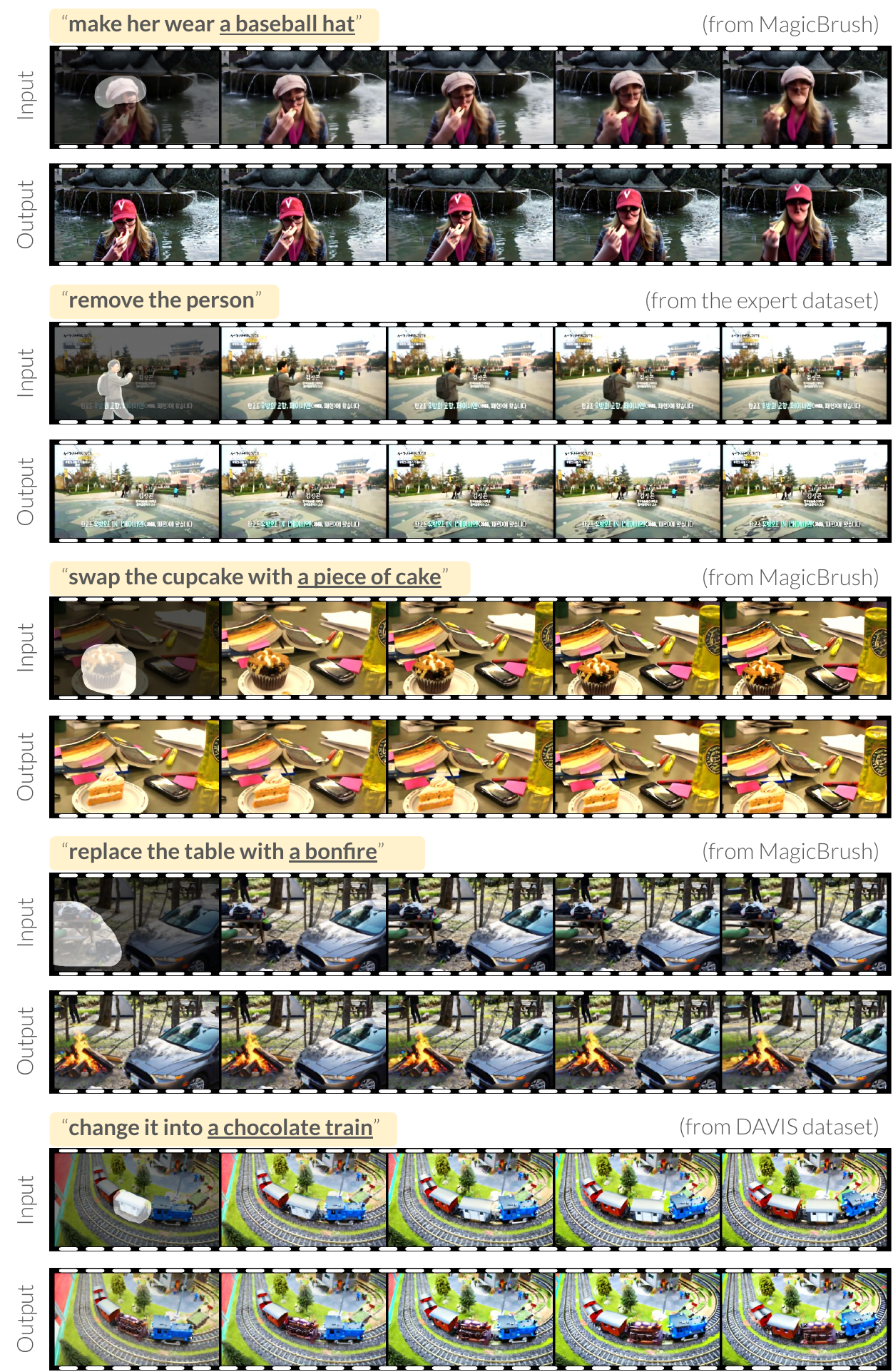}
  \caption{Examples of edited videos. Our method handles a diverse set of edit types; examples shown include appearance editing, object removal, semantic editing, and shape/texture editing. \methodnameshort\ successfully performs precise local edits while maintaining high visual fidelity to the source video.}
  \label{fig:examples}
\end{figure}

\begin{figure}[t!]
  \centering
  \includegraphics[width=\textwidth]{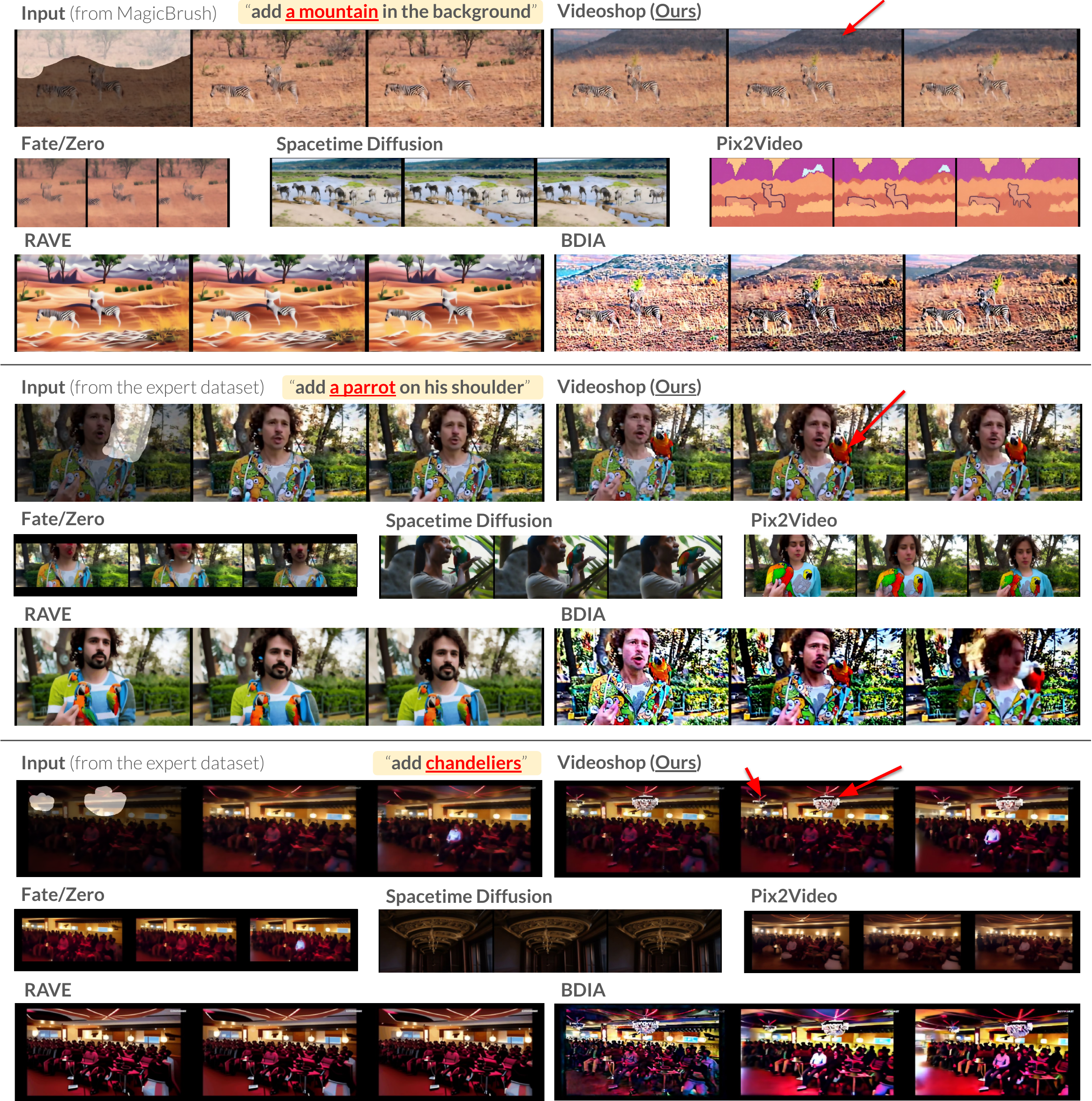}
  \caption{Qualitative comparison against baselines. \methodnameshort\ successfully maintains visual fidelity to the source video and target edit, while existing methods fail to do so.}
  \label{fig:vs_baselines}
  \vspace{-1.5em}
\end{figure}

\vspace{-.6em}
\subsubsection{Experimental setup.} We use the Stable Video Diffusion model\cite{blattmann2023stable} as the video generation model ($F_\theta$ in \cref{eq:edm_backward_with_noise_extrapolation}).
We compare our method against text-based video editing models including Pix2Video \cite{ceylan2023pix2video}, Fate/Zero\cite{qi2023fatezero}, Spacetime Diffusion \cite{yatim2023spacetime}, and RAVE \cite{kara2023rave}, as well as the exact inversion method BDIA \cite{zhang2023exact}. Pix2Video and RAVE are based on text-to-image models, while Fate/Zero and Spacetime Diffusion are based on text-to-video models. All baselines are implemented using their official codebases with videos resized to match the maximum supported video size. The baseline text-based video editing models either take a source and target prompt, or a single target prompt. To produce such source-target prompt pairs, we caption the first frame of the source video and update the modified concept to the target one (\eg, ``\texttt{a \underline{dog} running}'' $\rightarrow$ ``\texttt{a \underline{cat} running}'').

\vspace{-1.5em}
\subsubsection{Video editing capabilities of \methodnameshort.} We demonstrate the video editing capabilities of \methodnameshort\ in \cref{fig:teaser} and \cref{fig:examples}. We show a diverse example of video edits, including object addition (\cref{fig:teaser}(a)\circled{3}\circled{4}), object removal (\cref{fig:teaser}(c)\circled{2}, \cref{fig:examples}(b)), color edits (\cref{fig:teaser}(c)\circled{3}), semantic edits (\cref{fig:teaser}(c)\circled{1}, \cref{fig:examples}(c)(d)(e)), object attachment (\cref{fig:teaser}(b)\circled{2}) and appearance edits (\cref{fig:teaser}(a)\circled{1}, \cref{fig:examples}(a)). We also show examples of multiple edits at once (\cref{fig:teaser}). We observe that \methodnameshort\ successfully performs precise local edits, appearance control, and independent object addition, while maintaining high visual fidelity to the source video.

\vspace{-1em}
\subsubsection{Limitations of existing methods.} We compare \methodnameshort\ with existing methods in \cref{fig:vs_baselines}. We observe that existing methods often fail to maintain visual fidelity to the source video and target edit. For example, Fate/Zero does not correctly modify the source video based on the edit, Spacetime Diffusion generates videos that are structurally inconsistent with the source video, Pix2Video and RAVE can both undesirably change the style of the source video (\cref{fig:vs_baselines}(a)), and BDIA demonstrates large visual inconsistencies from the source video.

\vspace{-1em}
\subsubsection{Automated Evaluation.}
\label{subsec:automated_evaluation}

To assess generation quality, we adopt a comprehensive set of metrics. We use the terms ``source video frames,'' ``target image,'' and ``edited video frames'' to refer to the frames from the source video, the reference edited first frame, and the model-generated edited video frames. 
\begin{enumerate}
    \item \textit{Edit Fidelity}: Building on prior text-based video editing work~\cite{kara2023rave,yatim2023spacetime,ceylan2023pix2video}, we use the CLIP similarity~\cite{radford2021learning} metric. Our \textbf{\(\text{CLIP}_\text{tgt}\)} metric measures the similarity of CLIP embeddings between each edited frame and the target image. The \textbf{\(\text{CLIP}_\text{tgt}^+\)} score, utilizing CoTracker~\cite{karaev2023cotracker}, focuses only on the edited region. Furthermore, the \textbf{\(\text{TIFA}\)} score~\cite{hu2023tifa} evaluates semantic alignment between the target image and the edited video frames.
    
    \item \textit{Source Faithfulness}: We measure the \textbf{\(\text{CLIP}_\text{src}\)} similarity between the source and edited videos. To refine this, \textbf{\(\text{CLIP}_\text{src}^+\)} masks the edited region using CoTracker~\cite{karaev2023cotracker} and measures only the unedited region. Motion faithfulness is assessed via the end-point error (EPE) from optical flow comparisons using RAFT~\cite{teed2020raft}, which we denote as \textbf{\(\text{Flow}\)}. We also report the EPE within only unedited regions, denoted as \textbf{\(\text{Flow}^+\)}. \textbf{\(\text{FVD}\)} and \textbf{\(\text{SSIM}\)} scores provide additional quality measures.
    
    \item \textit{Temporal Consistency}: The average CLIP similarity between consecutive frames (\textbf{\(\text{CLIP}_\text{TC}\)}) is calculated following protocols from\cite{kara2023rave,ceylan2023pix2video,geyer2023tokenflow,yang2023rerender}.
\end{enumerate}

We report these metrics for both \methodnameshort\ and baseline methods, detailed in \cref{tab:quantitative_results} for MagicBrush and \cref{tab:quantitative_results_expert} for the expert dataset. In \cref{tab:quantitative_results}, {\methodnameshort} demonstrates superior performance over the baseline methods across the majority of metrics related to edit fidelity and source faithfulness, while showing competitive results in other evaluated areas, with marginal differences from the leading method. The findings in \cref{tab:quantitative_results_expert} echo this pattern, as {\methodnameshort} consistently ranks above the baselines in most metrics and holds competitive in the rest. Notably, the expert dataset exhibits an overall increase in optical flow errors, likely due to the more complex motion dynamics in real-world videos. Despite this, {\methodnameshort} maintains competitive scores on flow metrics and stands out particularly in preserving the source video flow in regions unoccupied by the edit.

\begin{table}[t]
  \caption{Quantitative results on MagicBrush. (\textBF{T.C.} = Temporal Consistency.)}
  \label{tab:quantitative_results}
  \vspace{-.9em}
  \centering
  \scriptsize
  \fontfamily{lmr2}\selectfont
  \begin{tabularx}{\textwidth}{l | *{3}{>{\centering\arraybackslash}X} @{\hspace{.6em}} | *{6}{>{\centering\arraybackslash}X} @{\hspace{.6em}} | *{1}{>{\centering\arraybackslash}X}}
    \toprule
    \multirow{3.5}{*}{\textBF{Method}} & \multicolumn{3}{c|}{\textBF{Edit Fidelity}} & \multicolumn{6}{c|}{\textBF{Source Faithfulness}} & \textBF{T.C.}\vspace{.2em} \\
    \cline{2-11}
    & \textBF{$\text{CLIP}_\text{tgt}$}$\uparrow$ & \textBF{$\text{CLIP}_{\text{tgt}}^+$}$\uparrow$ & \textBF{$\text{TIFA}$}$\uparrow$ & \textBF{$\text{CLIP}_\text{src}$}$\uparrow$ & \textBF{$\text{CLIP}_{\text{src}}^+$}$\uparrow$ & \textBF{$\text{Flow}$}$\downarrow$ & \textBF{$\text{Flow}^+$}$\downarrow$ & \textBF{FVD}$\downarrow$ & \textBF{SSIM}$\uparrow$ & \textBF{$\text{CLIP}_\text{TC}$}$\uparrow$ \\
    & ($\times 10^{-2}$) & ($\times 10^{-2}$) & ($\times 10^{-2}$) & ($\times 10^{-2}$) & ($\times 10^{-2}$) & ($\times 1$) & ($\times 1$) & ($\times 1$) & ($\times 10^{-2}$) & ($\times 10^{-2}$) \\
    \midrule
    \textBF{BDIA}\cite{zhang2023exact} & 82.12 & 82.19 & 57.67 & 82.48 & 87.10 & 2.83 & 1.43 & 3482.79 & 49.67 & 94.36 \\
    \midrule
    \textBF{Pix2Video}\cite{ceylan2023pix2video} & 71.19 & 76.47 & 51.98 & 74.55 & 79.03 & 3.59 & 2.58 & 2993.95 & 59.08 & 94.48 \\
    \textBF{Fate/Zero}\cite{qi2023fatezero} & 84.87 & 79.10 & 55.41 & \textBF{92.41} & 86.94 & 4.42 & 3.11 & 2205.03 & 48.59 & 95.71 \\
    \textBF{Spacetime}\cite{yatim2023spacetime} & 63.85 & 75.20 & 46.33 & 65.74 & 71.91 & 8.24 & 5.62 & 4815.63 & 41.61 & 96.58 \\
    \textBF{RAVE}\cite{kara2023rave} & 74.70 & 78.58 & 51.12 & 75.99 & 80.19 & 3.35 & 2.42 & 2354.09 & 62.21 & \textBF{96.59} \\
    \midrule
    \textBF{SVD} (no src. vid.) & 87.63 & 84.73 & 64.16 & 90.64 & 94.47 & 9.74 & 6.89 & 1894.16 & 47.50 & 95.07 \\
    \midrule
    \textBF{\methodnameshort} (Ours) & \textBF{88.80} & \textBF{85.58} & \textBF{64.40} & \underline{90.95} & \textBF{94.77} & \textBF{1.90} & \textBF{0.78} & \textBF{1478.76} & \textBF{71.92} & 95.16 \\
    \bottomrule
  \end{tabularx}
  \vspace{-1em}
\end{table}

\begin{table}[t]
  \caption{Quantitative results on the expert dataset. (\textBF{T.C.} = Temporal Consistency.)}
  \label{tab:quantitative_results_expert}
  \vspace{-.9em}
  \centering
  \scriptsize
  \fontfamily{lmr2}\selectfont
  \begin{tabularx}{\textwidth}{l | *{3}{>{\centering\arraybackslash}X} @{\hspace{.6em}} | *{6}{>{\centering\arraybackslash}X} @{\hspace{.6em}} | *{1}{>{\centering\arraybackslash}X}}
    \toprule
    \multirow{3.5}{*}{\textBF{Method}} & \multicolumn{3}{c|}{\textBF{Edit Fidelity}} & \multicolumn{6}{c|}{\textBF{Source Faithfulness}} & \textBF{T.C.}\vspace{.2em} \\
    \cline{2-11}
    & \textBF{$\text{CLIP}_\text{tgt}$}$\uparrow$ & \textBF{$\text{CLIP}_{\text{tgt}}^+$}$\uparrow$ & \textBF{$\text{TIFA}$}$\uparrow$ & \textBF{$\text{CLIP}_\text{src}$}$\uparrow$ & \textBF{$\text{CLIP}_{\text{src}}^+$}$\uparrow$ & \textBF{$\text{Flow}$}$\downarrow$ & \textBF{$\text{Flow}^+$}$\downarrow$ & \textBF{FVD}$\downarrow$ & \textBF{SSIM}$\uparrow$ & \textBF{$\text{CLIP}_\text{TC}$}$\uparrow$ \\
    & ($\times 10^{-2}$) & ($\times 10^{-2}$) & ($\times 10^{-2}$) & ($\times 10^{-2}$) & ($\times 10^{-2}$) & ($\times 1$) & ($\times 1$) & ($\times 1$) & ($\times 10^{-2}$) & ($\times 10^{-2}$) \\
    \midrule
    \textBF{BDIA}\cite{zhang2023exact} & 81.90 & 79.97 & 63.63 & 80.38 & 82.94 & 12.18 & 10.10 & 3048.53 & 36.42 & 92.86 \\
    \midrule
    \textBF{Pix2Video}\cite{ceylan2023pix2video} & 68.28 & 71.18 & 49.66 & 72.35 & 76.81 & 7.92 & 6.46 & 1876.43 & 61.23 & 93.90 \\
    \textBF{Fate/Zero}\cite{qi2023fatezero} & 71.51 & 72.53 & 48.65 & 74.59 & 82.68 & 9.70 & 8.30 & 2103.63 & 52.75 & 95.77 \\
    \textBF{Spacetime}\cite{yatim2023spacetime} & 58.61 & 68.45 & 43.06 & 56.51 & 65.86 & 7.93 & 6.31 & 4497.18 & 40.12 & \textBF{97.50} \\
    \textBF{RAVE}\cite{kara2023rave} & 71.59 & 71.39 & 57.78 & 74.37 & 76.35 & \textBF{5.59} & 4.54 & 1890.95 & 63.76 & 96.05 \\
    \midrule
    \textBF{\methodnameshort} (Ours) & \textBF{87.96} & \textBF{83.54} & \textBF{66.14} & \textBF{84.89} & \textBF{91.50} & \underline{5.85} & \textBF{4.47} & \textBF{1718.31} & \textBF{65.63} & 94.71 \\
    \bottomrule
  \end{tabularx}
  \vspace{-1.3em}
\end{table}

\vspace{-.6em}
\subsubsection{Human Evaluation.} We conduct a human evaluation study on the dev set of the MagicBrush dataset. For each baseline, we ask evaluators to compare the editing and video generation quality of our method with the baseline. The results are in the first two columns of \cref{tab:human_eval_and_time}. We observe that our method outperforms all baselines in both editing quality and video generation quality.

\vspace{-1em}

\subsubsection{Efficiency.} We assess the efficiency of \methodnameshort\ against baseline methods by measuring the average execution time per video, last column in \cref{tab:human_eval_and_time}. \methodnameshort\ aligns closely with the execution time of BDIA, which is known for its low overhead due to not requiring extra U-Net steps. Additionally, \methodnameshort\ provides a considerable speed advantage, operating at more than twice the speed (2.23x faster) of the average baseline method.

\definecolor{VChartColor1}{RGB}{60, 70, 255}
\definecolor{VChartColor2}{RGB}{237, 237, 237}
\newlength\VChartMax
\setlength\VChartMax{4.5em}
\newcommand*\VChart[3]{~\rlap{\textcolor{VChartColor2}{\rule{1\VChartMax}{1ex}}}\rlap{\textcolor{VChartColor2}{\rule{#3\VChartMax}{1ex}}}\rlap{\textcolor{VChartColor1}{\rule{#2\VChartMax}{1ex}}}\hphantom{\rule{1\VChartMax}{1ex}}~~~#1\%}
\begin{table}[t]
  \caption{Human evaluation and execution time. For human evaluation, evaluators are asked to compare the editing quality and video generation quality of our method against each baseline. \methodnameshort\ outperforms all baselines in both editing quality and video generation quality and has a competitive runtime compared to the baseline methods.}
  \vspace{-.8em}
  \label{tab:human_eval_and_time}
  \centering
  \scriptsize
  \fontfamily{lmr2}\selectfont
  \begin{tabularx}{.9\textwidth}{l | *{3}{>{\centering\arraybackslash}X}}
    \toprule
    \textBF{\methodnameshort\ (Ours)} & \textBF{Editing Quality} & \textBF{Video Generation Quality} & \textBF{Execution Time} \\
    \textBF{vs. ...} & (preference in our favor \%) & (preference in our favor \%) & (as multiples of ours) \\
    \midrule
    \textBF{BDIA}\cite{zhang2023exact} & \VChart{94.89}{.9489}{1} & \VChart{90.53}{.9053}{1} & 1.03x \\
    \midrule
    \textBF{Pix2Video}\cite{ceylan2023pix2video} & \VChart{98.30}{.9830}{1} & \VChart{96.97}{.9697}{1} & 4.70x \\
    \textBF{Fate/Zero}\cite{qi2023fatezero} & \VChart{98.67}{.9867}{1} & \VChart{92.23}{.9223}{1} & 0.71x \\
    \textBF{Spacetime}\cite{yatim2023spacetime} & \VChart{99.81}{.9981}{1} & \VChart{69.32}{.6932}{1} & 3.41x \\
    \textBF{RAVE}\cite{kara2023rave} & \VChart{92.99}{.9299}{1} & \VChart{74.05}{.7405}{1} & 1.31x \\
    \midrule
    \textit{\textBF{Average}} & \VChart{96.93}{.9693}{1} & \VChart{84.62}{.8462}{1} & 2.23x \\
    \bottomrule
  \end{tabularx}
  \vspace{-1em}
\end{table}

\vspace{-.8em}
\subsection{Ablation Study}
\label{sec:ablation}
\vspace{-.2em}

\begin{figure}[t]
  \centering
  \includegraphics[width=\textwidth]{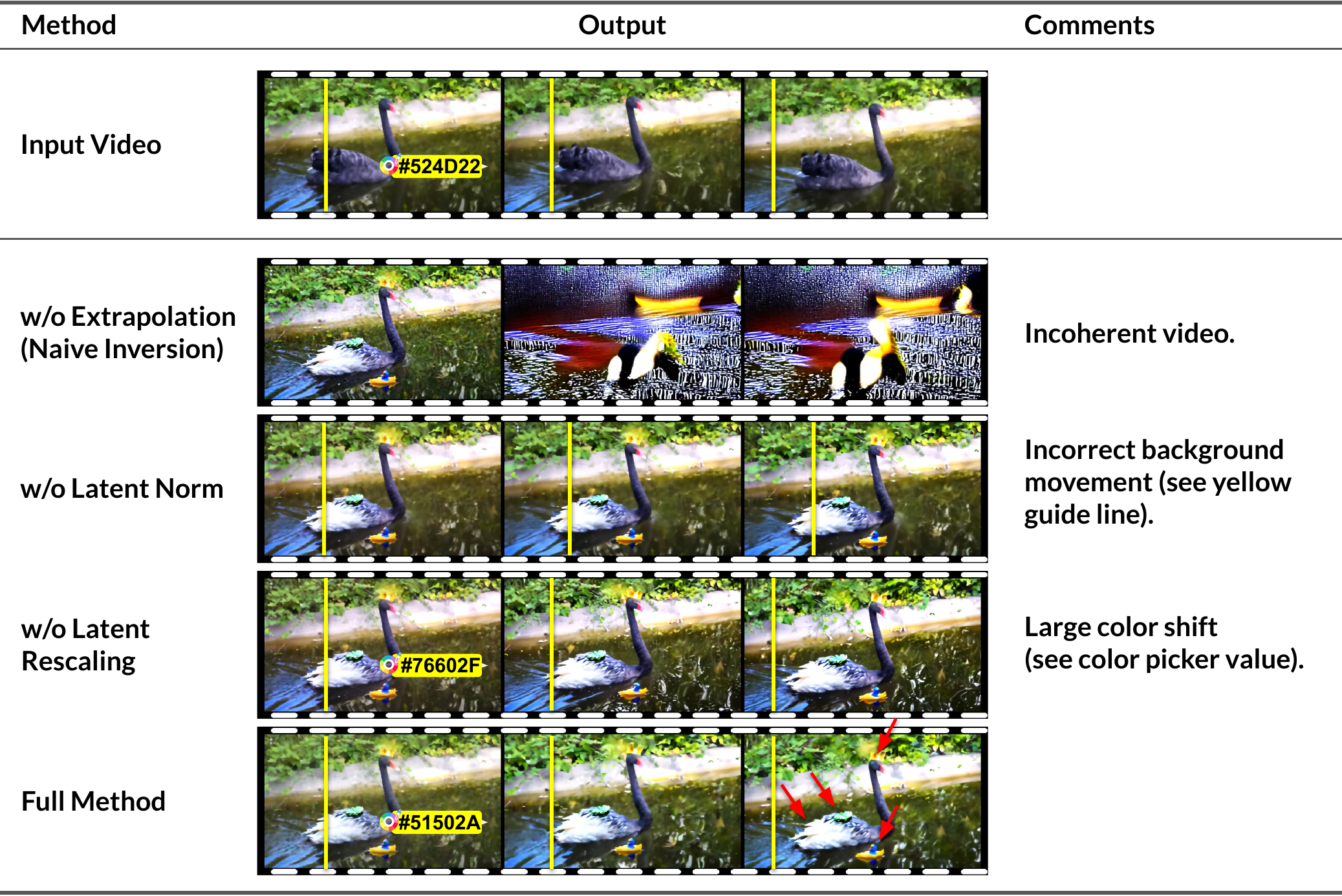}
  \vspace{-2em}
  \caption{Examples from ablations. The yellow vertical line tracks the background movement and the yellow color picker shows the RGB value of a background pixel at the same location in the first frame. As we can see from the examples, the lack of noise extrapolation results in incoherent videos, the lack of latent normalization results in a slow, incorrect background movement that disregards the source video's movement, and the lack of latent rescaling results in shifted colors.}
  \label{fig:ablation_examples}
  \vspace{-.6em}
\end{figure}

We ablate latent normalization (\cref{sec:latent_normalization}), latent rescaling (\cref{sec:latent_normalization}), noise extrapolation (\cref{sec:noise_extrapolation}), and noise threshold (\cref{sec:noise_extrapolation}) qualitatively in \cref{fig:ablation_examples} and quantitatively in \cref{tab:ablations}. In \cref{fig:ablation_examples}, the yellow vertical line tracks the background movement and the yellow color picker shows the RGB value of a background pixel at the same location in the first frame. As we can see, the lack of noise extrapolation results in incoherent videos, the lack of latent normalization results in a slow, incorrect background movement that disregards motion in the source video, and removing latent rescaling results in shifted colors. In \cref{tab:ablations}, we observe that our final method outperforms all ablations except for a few scores without latent normalization. However, it is important to note that the lack of latent normalization tends to result in static or slow movements that disregard motion in the source video, and therefore are unsuitable for video editing. Furthermore, removing the threshold in noise extrapolation leads to division by a very small number, resulting in numerical instability and NaN values in the latents.

\begin{table}[t]
  \caption{Ablations on a 10-video subset. Our method outperforms all ablations except for a few scores when removing latent norm (second row). However, it is important to note that the lack of latent norm tends to result in static or slow movements that disregard the movement in the source video, as evidenced by the low Flow scores. Furthermore, removing the noise threshold (last row) leads to division by a very small number, resulting in NaN values in the latents. (\textBF{T.C.} = Temporal Consistency.)}
  \vspace{-.6em}
  \label{tab:ablations}
  \centering
  \scriptsize
  \fontfamily{lmr2}\selectfont
  \begin{tabularx}{\textwidth}{l | *{3}{>{\centering\arraybackslash}X} @{\hspace{.6em}} | *{6}{>{\centering\arraybackslash}X} @{\hspace{.6em}} | *{1}{>{\centering\arraybackslash}X}}
    \toprule
    \multirow{3.5}{*}{\textBF{Method}} & \multicolumn{3}{c|}{\textBF{Edit Fidelity}} & \multicolumn{6}{c|}{\textBF{Source Faithfulness}} & \textBF{T.C.}\vspace{.2em} \\
    \cline{2-11}
    & \textBF{$\text{CLIP}_\text{tgt}$}$\uparrow$ & \textBF{$\text{CLIP}_{\text{tgt}}^+$}$\uparrow$ & \textBF{$\text{TIFA}$}$\uparrow$ & \textBF{$\text{CLIP}_\text{src}$}$\uparrow$ & \textBF{$\text{CLIP}_{\text{src}}^+$}$\uparrow$ & \textBF{$\text{Flow}$}$\downarrow$ & \textBF{$\text{Flow}^+$}$\downarrow$ & \textBF{FVD}$\downarrow$ & \textBF{SSIM}$\uparrow$ & \textBF{$\text{CLIP}_\text{TC}$}$\uparrow$ \\
    & ($\times 10^{-2}$) & ($\times 10^{-2}$) & ($\times 10^{-2}$) & ($\times 10^{-2}$) & ($\times 10^{-2}$) & ($\times 1$) & ($\times 1$) & ($\times 1$) & ($\times 10^{-2}$) & ($\times 10^{-2}$) \\
    \midrule
    \textBF{\methodnameshort} (Ours) & \textBF{92.12} & \underline{88.89} & \textBF{100} & \underline{88.82} & \textBF{97.92} & \textBF{4.57} & \textBF{2.42} & \underline{1410.70} & \textBF{57.25} & \underline{95.18} \\
    \textBF{-- Norm.} & 91.36 & \textBF{91.54} & \textBF{100} & \textBF{90.22} & \textBF{97.34} & \textBF{\color{red} 19.57} & \textBF{\color{red} 13.00} & \textBF{1324.45} & 41.74 & \textBF{96.58} \\
    \textBF{-- Rescaling} & 89.74 & 88.45 & 90 & 88.72 & 97.34 & 6.37 & 3.64 & 1784.10 & 52.52 & 94.60 \\
    \textBF{-- Extrapolation} & 73.36 & 75.42 & 70 & 87.19 & 95.12 & 17.92 & 10.47 & 5316.65 & 16.76 & 84.55 \\
    \textBF{-- Threshold} & NaN & NaN & NaN & NaN & NaN & NaN & NaN & NaN & NaN & NaN \\
    \bottomrule
  \end{tabularx}
  \vspace{-2.5em}
\end{table}

\vspace{-.8em}
\subsection{Extension to Other Models and Longer Videos}
\vspace{-.2em}

\begin{figure}[t]
    \centering
    \begin{subfigure}[t]{0.4\textwidth}
        \centering
        \begin{minipage}[c]{0.25\linewidth}
            \includegraphics[width=\linewidth]{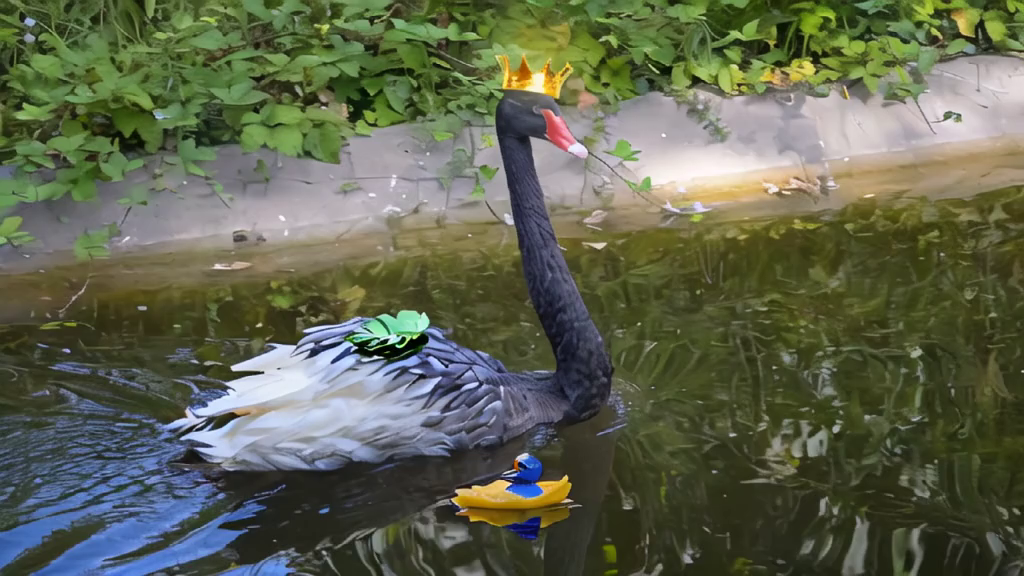}
            \caption*{\scriptsize Frame 1}
        \end{minipage}
        \begin{minipage}[c]{0.25\linewidth}
            \includegraphics[width=\linewidth]{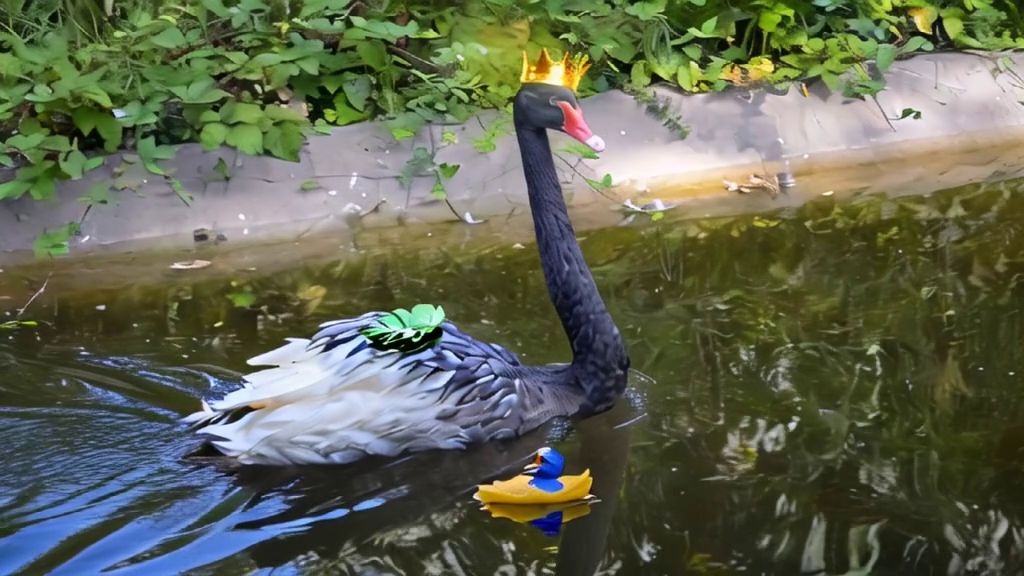}
            \caption*{\scriptsize Frame 7}
        \end{minipage}
        \begin{minipage}[c]{0.25\linewidth}
            \includegraphics[width=\linewidth]{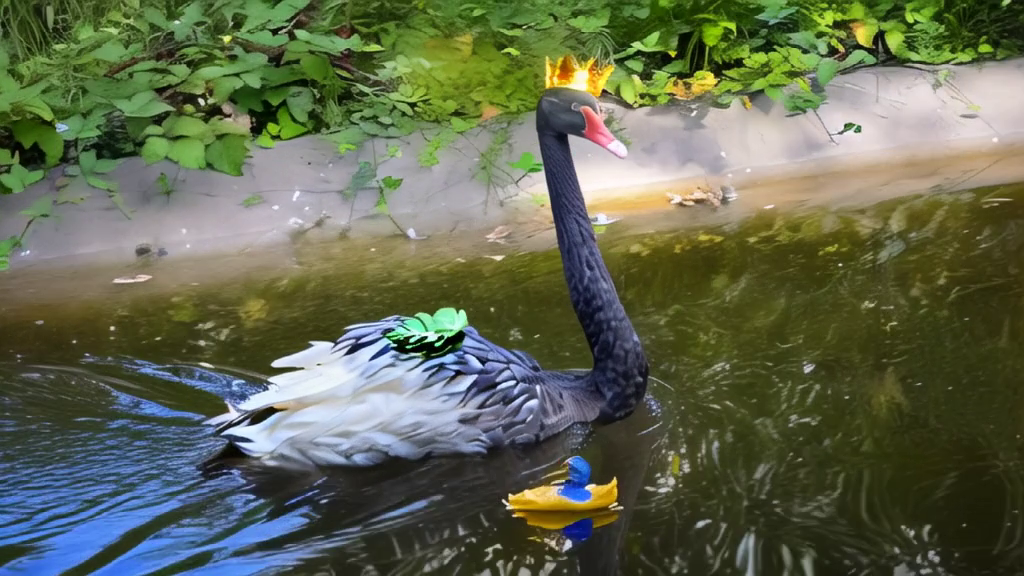}
            \caption*{\scriptsize Frame 14}
        \end{minipage}
        \caption{Results on AnimateLCM~\cite{wang2024animatelcm}.}
        \label{fig:results_animatelcm}
    \end{subfigure}%
    ~
    \begin{subfigure}[t]{0.4\textwidth}
        \centering
        \begin{minipage}[c]{0.25\linewidth}
            \includegraphics[width=\linewidth]{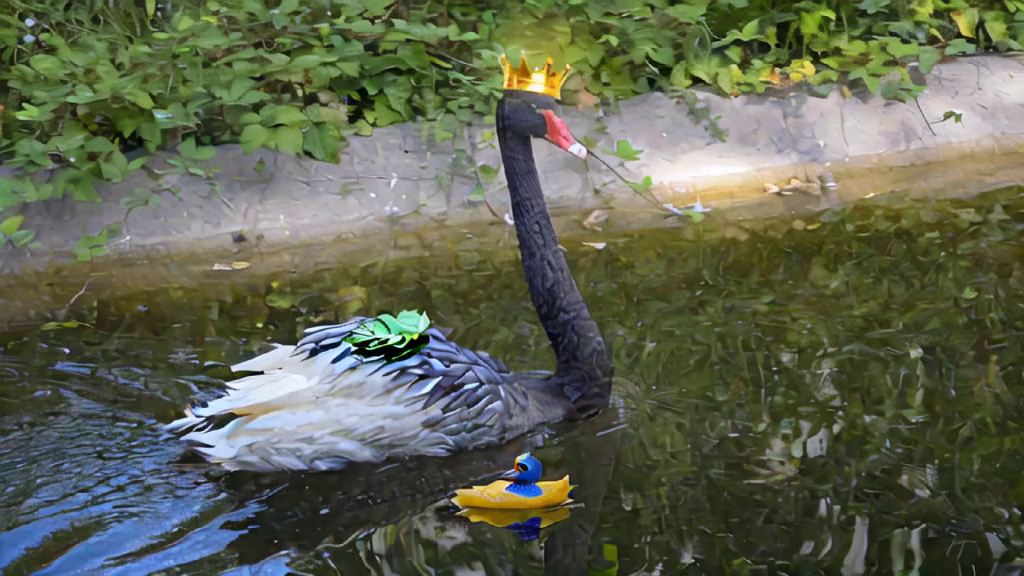}
            \caption*{\scriptsize Frame 1}
        \end{minipage}
        \begin{minipage}[c]{0.25\linewidth}
            \includegraphics[width=\linewidth]{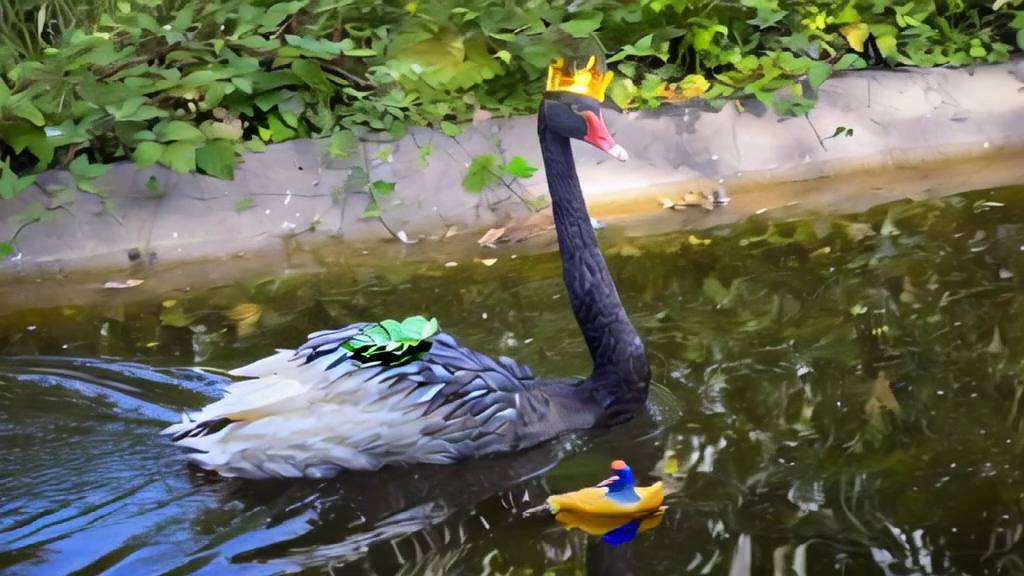}
            \caption*{\scriptsize Frame 25}
        \end{minipage}
        \begin{minipage}[c]{0.25\linewidth}
            \includegraphics[width=\linewidth]{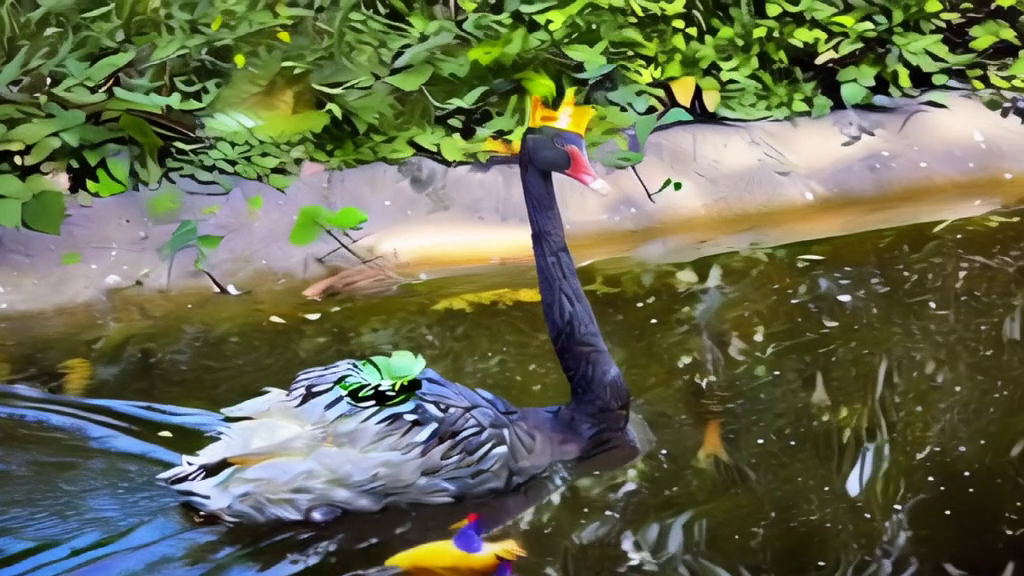}
            \caption*{\scriptsize Frame 50}
        \end{minipage}
        \caption{Expanded frames (50 total).}
        \label{fig:expanded_frames}
    \end{subfigure}
    \vspace{-.8em}
    \caption{Our method can extend to other video diffusion models and longer videos.}
    \vspace{-2em}
\end{figure}

Our approach can be applied to any image-to-video diffusion model with an invertible denoising step. We show results on the AnimateLCM~\cite{wang2024animatelcm} model, a recent image-to-video model, in \cref{fig:results_animatelcm}.
The number of frames in our method can be naturally extended by recurrently generating blocks of frames, each block conditioned on the last frame of the previous block as shown in our \cref{fig:expanded_frames}.

\vspace{-.65em}
\section{Discussion and Limitations}
\vspace{-.25em}
\methodname\ is a novel video editing approach that lets users make localized semantic edits without any need for re-training. Other video editing methods currently edit whole videos with sparse textual instructions. Instead, we simplify the problem by reducing it to image editing, a well-studied and widely applied task in the image domain. We exploited the linear relationship in the latents during the inversion process with our noise extrapolation and showed our latent normalization and rescaling enables realistic localized semantic editing.   

Despite the strengths of our method, there are limitations: (a) The VAE used in our method may lead to loss of information during video encoding, which could obscure some fine details such as small text. (b) In cases where the source video contains large movements or flickering, the temporal consistency of our method may be compromised. (c) Our method is training-free, which means that it only uses the knowledge contained in the base model. This limits the introduction of new information (such as new motion). We anticipate that our approach will become even more effective as image-to-video models improve. Finally, we can leverage video models to extend our method for editing 3D meshes~\cite{decatur20233d, decatur20243d} and extracting visual knowledge by analyzing the latent~\cite{du2023generative, sarkar2024shadows, bhattad2024stylegan}. Another line of work includes combining image editing with motion and trajectory controls to ensure seamless video results. In summary, \methodname\ reimagines video editing and could unlock possibilities for new and creative applications. With our method, users can edit videos with the same ease as they would edit images in Photoshop.  

\vspace{0.5em}
\noindent\textbf{Acknowledgement.} This work was supported by NSF award IIS-2211133.

\newpage
\bibliographystyle{splncs04}
\bibliography{references}

\begin{thebibliography}{10}
\providecommand{\url}[1]{\texttt{#1}}
\providecommand{\urlprefix}{URL }
\providecommand{\doi}[1]{https://doi.org/#1}

\bibitem{bhattad2022cut}
Bhattad, A., Forsyth, D.A.: Cut-and-paste object insertion by enabling deep image prior for reshading. In: 2022 International Conference on 3D Vision (3DV). pp. 332--341. IEEE (2022)

\bibitem{bhattad2024stylegan}
Bhattad, A., McKee, D., Hoiem, D., Forsyth, D.: Stylegan knows normal, depth, albedo, and more. Advances in Neural Information Processing Systems  \textbf{36} (2024)

\bibitem{bhattad2024stylitgan}
Bhattad, A., Soole, J., Forsyth, D.: Stylitgan: Image-based relighting via latent control. In: Proceedings of the IEEE/CVF Conference on Computer Vision and Pattern Recognition. pp. 4231--4240 (2024)

\bibitem{blattmann2023stable}
Blattmann, A., Dockhorn, T., Kulal, S., Mendelevitch, D., Kilian, M., Lorenz, D., Levi, Y., English, Z., Voleti, V., Letts, A., Jampani, V., Rombach, R.: Stable video diffusion: Scaling latent video diffusion models to large datasets (2023)

\bibitem{blattmann2023align}
Blattmann, A., Rombach, R., Ling, H., Dockhorn, T., Kim, S.W., Fidler, S., Kreis, K.: Align your latents: High-resolution video synthesis with latent diffusion models. In: Proceedings of the IEEE/CVF Conference on Computer Vision and Pattern Recognition. pp. 22563--22575 (2023)

\bibitem{ceylan2023pix2video}
Ceylan, D., Huang, C.H.P., Mitra, N.J.: Pix2video: Video editing using image diffusion (2023)

\bibitem{chan2019everybody}
Chan, C., Ginosar, S., Zhou, T., Efros, A.A.: Everybody dance now (2019)

\bibitem{chang2023diffusionatlas}
Chang, S.Y., Chen, H.T., Liu, T.L.: Diffusionatlas: High-fidelity consistent diffusion video editing (2023)

\bibitem{chen2024videocrafter2}
Chen, H., Zhang, Y., Cun, X., Xia, M., Wang, X., Weng, C., Shan, Y.: Videocrafter2: Overcoming data limitations for high-quality video diffusion models (2024)

\bibitem{chen2023anydoor}
Chen, X., Huang, L., Liu, Y., Shen, Y., Zhao, D., Zhao, H.: Anydoor: Zero-shot object-level image customization. arXiv preprint arXiv:2307.09481  (2023)

\bibitem{couairon2023videdit}
Couairon, P., Rambour, C., Haugeard, J.E., Thome, N.: Videdit: Zero-shot and spatially aware text-driven video editing (2023)

\bibitem{decatur20243d}
Decatur, D., Lang, I., Aberman, K., Hanocka, R.: 3d paintbrush: Local stylization of 3d shapes with cascaded score distillation. arXiv preprint arXiv:2311.09571  (2023)

\bibitem{decatur20233d}
Decatur, D., Lang, I., Hanocka, R.: 3d highlighter: Localizing regions on 3d shapes via text descriptions. In: Proceedings of the IEEE/CVF Conference on Computer Vision and Pattern Recognition. pp. 20930--20939 (2023)

\bibitem{du2023generative}
Du, X., Kolkin, N., Shakhnarovich, G., Bhattad, A.: Generative models: What do they know? do they know things? let's find out! arXiv preprint arXiv:2311.17137  (2023)

\bibitem{gal2022image}
Gal, R., Alaluf, Y., Atzmon, Y., Patashnik, O., Bermano, A.H., Chechik, G., Cohen-Or, D.: An image is worth one word: Personalizing text-to-image generation using textual inversion. arXiv preprint arXiv:2208.01618  (2022)

\bibitem{geyer2023tokenflow}
Geyer, M., Bar-Tal, O., Bagon, S., Dekel, T.: Tokenflow: Consistent diffusion features for consistent video editing (2023)

\bibitem{goel2023pair}
Goel, V., Peruzzo, E., Jiang, Y., Xu, D., Sebe, N., Darrell, T., Wang, Z., Shi, H.: Pair-diffusion: Object-level image editing with structure-and-appearance paired diffusion models. arXiv preprint arXiv:2303.17546  (2023)

\bibitem{goodfellow2014generative}
Goodfellow, I., Pouget-Abadie, J., Mirza, M., Xu, B., Warde-Farley, D., Ozair, S., Courville, A., Bengio, Y.: Generative adversarial nets. Advances in neural information processing systems  \textbf{27} (2014)

\bibitem{gutmann2010noise}
Gutmann, M., Hyv{\"a}rinen, A.: Noise-contrastive estimation: A new estimation principle for unnormalized statistical models. In: Proceedings of the thirteenth international conference on artificial intelligence and statistics. pp. 297--304. JMLR Workshop and Conference Proceedings (2010)

\bibitem{hertz2022prompt}
Hertz, A., Mokady, R., Tenenbaum, J., Aberman, K., Pritch, Y., Cohen-Or, D.: Prompt-to-prompt image editing with cross attention control. arXiv preprint arXiv:2208.01626  (2022)

\bibitem{ho2020denoising}
Ho, J., Jain, A., Abbeel, P.: Denoising diffusion probabilistic models (2020)

\bibitem{hu2023tifa}
Hu, Y., Liu, B., Kasai, J., Wang, Y., Ostendorf, M., Krishna, R., Smith, N.A.: Tifa: Accurate and interpretable text-to-image faithfulness evaluation with question answering (2023)

\bibitem{AppleFinalCutPro}
Inc., A.: Final cut pro. \url{https://www.apple.com/final-cut-pro/} (2023), accessed: 2024-03-03

\bibitem{AdobePremierePro}
Incorporated, A.S.: Adobe premiere pro. \url{https://www.adobe.com/products/premiere.html} (2023), accessed: 2024-03-03

\bibitem{jeong2024groundavideo}
Jeong, H., Ye, J.C.: Ground-a-video: Zero-shot grounded video editing using text-to-image diffusion models (2024)

\bibitem{kahatapitiya2024objectcentric}
Kahatapitiya, K., Karjauv, A., Abati, D., Porikli, F., Asano, Y.M., Habibian, A.: Object-centric diffusion for efficient video editing (2024)

\bibitem{kang2023scaling}
Kang, M., Zhu, J.Y., Zhang, R., Park, J., Shechtman, E., Paris, S., Park, T.: Scaling up gans for text-to-image synthesis. In: Proceedings of the IEEE/CVF Conference on Computer Vision and Pattern Recognition. pp. 10124--10134 (2023)

\bibitem{kara2023rave}
Kara, O., Kurtkaya, B., Yesiltepe, H., Rehg, J.M., Yanardag, P.: Rave: Randomized noise shuffling for fast and consistent video editing with diffusion models (2023)

\bibitem{karaev2023cotracker}
Karaev, N., Rocco, I., Graham, B., Neverova, N., Vedaldi, A., Rupprecht, C.: Cotracker: It is better to track together (2023)

\bibitem{karras2022elucidating}
Karras, T., Aittala, M., Aila, T., Laine, S.: Elucidating the design space of diffusion-based generative models (2022)

\bibitem{karras2019style}
Karras, T., Laine, S., Aila, T.: A style-based generator architecture for generative adversarial networks. In: Proceedings of the IEEE/CVF conference on computer vision and pattern recognition. pp. 4401--4410 (2019)

\bibitem{kasten2021layered}
Kasten, Y., Ofri, D., Wang, O., Dekel, T.: Layered neural atlases for consistent video editing (2021)

\bibitem{kondratyuk2023videopoet}
Kondratyuk, D., Yu, L., Gu, X., Lezama, J., Huang, J., Hornung, R., Adam, H., Akbari, H., Alon, Y., Birodkar, V., et~al.: Videopoet: A large language model for zero-shot video generation. arXiv preprint arXiv:2312.14125  (2023)

\bibitem{mackay1994video}
Mackay, W., Pagani, D.: Video mosaic: Laying out time in a physical space. In: Proceedings of the second ACM international conference on Multimedia. pp. 165--172 (1994)

\bibitem{meiri2023fixedpoint}
Meiri, B., Samuel, D., Darshan, N., Chechik, G., Avidan, S., Ben-Ari, R.: Fixed-point inversion for text-to-image diffusion models (2023)

\bibitem{menapace2024snap}
Menapace, W., Siarohin, A., Skorokhodov, I., Deyneka, E., Chen, T.S., Kag, A., Fang, Y., Stoliar, A., Ricci, E., Ren, J., et~al.: Snap video: Scaled spatiotemporal transformers for text-to-video synthesis. arXiv preprint arXiv:2402.14797  (2024)

\bibitem{michel2024object}
Michel, O., Bhattad, A., VanderBilt, E., Krishna, R., Kembhavi, A., Gupta, T.: Object 3dit: Language-guided 3d-aware image editing. Advances in Neural Information Processing Systems  \textbf{36} (2024)

\bibitem{mokady2022nulltext}
Mokady, R., Hertz, A., Aberman, K., Pritch, Y., Cohen-Or, D.: Null-text inversion for editing real images using guided diffusion models (2022)

\bibitem{Mullan_Hotshot-XL_2023}
Mullan, J., Crawbuck, D., Sastry, A.: {Hotshot-XL} (Oct 2023), \url{https://github.com/hotshotco/hotshot-xl}

\bibitem{qi2023fatezero}
Qi, C., Cun, X., Zhang, Y., Lei, C., Wang, X., Shan, Y., Chen, Q.: Fatezero: Fusing attentions for zero-shot text-based video editing (2023)

\bibitem{radford2021learning}
Radford, A., Kim, J.W., Hallacy, C., Ramesh, A., Goh, G., Agarwal, S., Sastry, G., Askell, A., Mishkin, P., Clark, J., Krueger, G., Sutskever, I.: Learning transferable visual models from natural language supervision (2021)

\bibitem{ren2024customizeavideo}
Ren, Y., Zhou, Y., Yang, J., Shi, J., Liu, D., Liu, F., Kwon, M., Shrivastava, A.: Customize-a-video: One-shot motion customization of text-to-video diffusion models (2024)

\bibitem{rombach2022highresolution}
Rombach, R., Blattmann, A., Lorenz, D., Esser, P., Ommer, B.: High-resolution image synthesis with latent diffusion models (2022)

\bibitem{santosa2013direct}
Santosa, S., Chevalier, F., Balakrishnan, R., Singh, K.: Direct space-time trajectory control for visual media editing. In: Proceedings of the SIGCHI Conference on Human Factors in Computing Systems. pp. 1149--1158 (2013)

\bibitem{sarkar2024shadows}
Sarkar, A., Mai, H., Mahapatra, A., Lazebnik, S., Forsyth, D.A., Bhattad, A.: Shadows don't lie and lines can't bend! generative models don't know projective geometry... for now. In: Proceedings of the IEEE/CVF Conference on Computer Vision and Pattern Recognition. pp. 28140--28149 (2024)

\bibitem{sauer2023stylegan}
Sauer, A., Karras, T., Laine, S., Geiger, A., Aila, T.: Stylegan-t: Unlocking the power of gans for fast large-scale text-to-image synthesis. arXiv preprint arXiv:2301.09515  (2023)

\bibitem{shin2023editavideo}
Shin, C., Kim, H., Lee, C.H., gil Lee, S., Yoon, S.: Edit-a-video: Single video editing with object-aware consistency (2023)

\bibitem{singer2022make}
Singer, U., Polyak, A., Hayes, T., Yin, X., An, J., Zhang, S., Hu, Q., Yang, H., Ashual, O., Gafni, O., et~al.: Make-a-video: Text-to-video generation without text-video data. arXiv preprint arXiv:2209.14792  (2022)

\bibitem{song2022denoising}
Song, J., Meng, C., Ermon, S.: Denoising diffusion implicit models (2022)

\bibitem{teed2020raft}
Teed, Z., Deng, J.: Raft: Recurrent all-pairs field transforms for optical flow (2020)

\bibitem{vincent2011connection}
Vincent, P.: A connection between score matching and denoising autoencoders. Neural computation  \textbf{23}(7),  1661--1674 (2011)

\bibitem{wallace2022edict}
Wallace, B., Gokul, A., Naik, N.: Edict: Exact diffusion inversion via coupled transformations (2022)

\bibitem{wang2024animatelcm}
Wang, F.Y., Huang, Z., Shi, X., Bian, W., Song, G., Liu, Y., Li, H.: Animatelcm: Accelerating the animation of personalized diffusion models and adapters with decoupled consistency learning (2024)

\bibitem{wang2023modelscope}
Wang, J., Yuan, H., Chen, D., Zhang, Y., Wang, X., Zhang, S.: Modelscope text-to-video technical report (2023)

\bibitem{wang2023videocomposer}
Wang, X., Yuan, H., Zhang, S., Chen, D., Wang, J., Zhang, Y., Shen, Y., Zhao, D., Zhou, J.: Videocomposer: Compositional video synthesis with motion controllability (2023)

\bibitem{xue2022advancing}
Xue, H., Hang, T., Zeng, Y., Sun, Y., Liu, B., Yang, H., Fu, J., Guo, B.: Advancing high-resolution video-language representation with large-scale video transcriptions (2022)

\bibitem{yang2023rerender}
Yang, S., Zhou, Y., Liu, Z., Loy, C.C.: Rerender a video: Zero-shot text-guided video-to-video translation (2023)

\bibitem{yatim2023spacetime}
Yatim, D., Fridman, R., Bar-Tal, O., Kasten, Y., Dekel, T.: Space-time diffusion features for zero-shot text-driven motion transfer (2023)

\bibitem{yenphraphai2024image}
Yenphraphai, J., Pan, X., Liu, S., Panozzo, D., Xie, S.: Image sculpting: Precise object editing with 3d geometry control. arXiv preprint arXiv:2401.01702  (2024)

\bibitem{yin2023dance}
Yin, W., Yin, H., Baraka, K., Kragic, D., Björkman, M.: Dance style transfer with cross-modal transformer (2023)

\bibitem{zhang2023show1}
Zhang, D.J., Wu, J.Z., Liu, J.W., Zhao, R., Ran, L., Gu, Y., Gao, D., Shou, M.Z.: Show-1: Marrying pixel and latent diffusion models for text-to-video generation (2023)

\bibitem{zhang2023exact}
Zhang, G., Lewis, J.P., Kleijn, W.B.: Exact diffusion inversion via bi-directional integration approximation (2023)

\bibitem{zhang2023magicbrush}
Zhang, K., Mo, L., Chen, W., Sun, H., Su, Y.: Magicbrush: A manually annotated dataset for instruction-guided image editing (2023)

\bibitem{zhang2023adding}
Zhang, L., Rao, A., Agrawala, M.: Adding conditional control to text-to-image diffusion models. In: Proceedings of the IEEE/CVF International Conference on Computer Vision. pp. 3836--3847 (2023)

\bibitem{zuo2023cutandpaste}
Zuo, Z., Zhang, Z., Luo, Y., Zhao, Y., Zhang, H., Yang, Y., Wang, M.: Cut-and-paste: Subject-driven video editing with attention control (2023)

\end{thebibliography}

\end{document}